%% file: main.tex
\documentclass[twoside]{article}

\usepackage[accepted]{aistats2024arxiv}

\usepackage[round]{natbib}
\renewcommand{\bibname}{References}
\renewcommand\refname{References}
\renewcommand{\bibsection}{\subsubsection*{\bibname}}

\usepackage[utf8]{inputenc} %
\usepackage[T1]{fontenc}    %

\usepackage{booktabs}       %
\usepackage{amsfonts}       %
\usepackage{nicefrac}       %
\usepackage{microtype}      %

\usepackage{graphicx}

\usepackage{url}
\usepackage[symbol]{footmisc}

\usepackage{todonotes}
\usepackage{algorithm,algorithmic}
\usepackage{amsmath}
\usepackage{amssymb}
\usepackage{caption}
\usepackage{subcaption}
\usepackage{amsthm}
\usepackage{bbm}
\usepackage{array,multirow}
\usepackage{wrapfig}
\usepackage[normalem]{ulem}
\usepackage{makecell}
\usepackage{pifont}

\usepackage[english]{babel}
\addto{\captionsenglish}{%
    \renewcommand{\bibname}{References}
}
\usepackage{mdframed}   %

\definecolor{mydarkblue}{rgb}{0,0.08,0.45}
\usepackage[colorlinks,citecolor=mydarkblue,urlcolor=mydarkblue,linkcolor=mydarkblue,filecolor=mydarkblue]{hyperref}

\input{math_commands.tex}
\theoremstyle{plain}

\theoremstyle{definition}

\theoremstyle{remark}

\usepackage[capitalize]{cleveref}
\crefname{appendix}{App.}{Apps.}

\begin{document}

\twocolumn[

\aistatstitle{Learning to Defer to a Population: A Meta-Learning Approach}

\runningauthor{D.\ Tailor, A.\ Patra, R.\ Verma, P.\ Manggala, E.\ Nalisnick}

\aistatsauthor{
Dharmesh Tailor ~
Aditya Patra ~
Rajeev Verma ~
Putra Manggala ~
Eric Nalisnick ~
}
\aistatsaddress{University of Amsterdam} ]

\begin{abstract}
The \textit{learning to defer} (L2D) framework allows autonomous systems to be safe and robust by allocating difficult decisions to a human expert. All existing work on L2D assumes that each expert is well-identified, and if any expert were to change, the system should be re-trained.  In this work, we alleviate this constraint, formulating an L2D system that can cope with never-before-seen experts at test-time.  We accomplish this by using \textit{meta-learning}, considering both optimization- and model-based variants.  Given a small context set to characterize the currently available expert, our framework can quickly adapt its deferral policy.  For the model-based approach, we employ an attention mechanism that is able to look for points in the context set that are similar to a given test point, leading to an even more precise assessment of the expert's abilities.  In the experiments, we validate our methods on image recognition, traffic sign detection, and skin lesion diagnosis benchmarks.
\end{abstract}

\section{INTRODUCTION}
\label{sec:intro}
\textit{Hybrid intelligent} (HI) systems \citep{kamar2016directions, dellermann2019hybrid, akata2020research} assume some form of on-going collaboration between humans and machines.  While this can take many forms, our work focuses on the paradigm of \textit{learning to defer} (L2D) \citep{madras2018predict}: HI systems that can defer to a human upon facing challenging or high-risk decisions.  An instructive example is that of automated medicine.  Given patient data, the HI system can either make a diagnosis---if it is confident in its prediction---or call in a human doctor to take the case. 

Current L2D systems are trained so that the model is customized to one \citep{pmlr-v119-mozannar20b} or more \citep{verma2023learning} specific humans.  If the experts'\footnote{We use the terms \textit{human} and \textit{expert} interchangeably, since we assume that all humans involved can possibly outperform the predictive model.} behavior changes from training- to test-time, the system will break, mis-allocating instances to the human when the machine would perform better or vice versa.  An extreme form of this distribution shift is when the expert's identity completely changes.  Returning to the medical setting, this means that when one doctor leaves duty and another takes her place, an \emph{entirely} new L2D system must be brought online.  Similarly, when a new doctor joins the staff, a new L2D system must be trained from scratch (although one can imagine ways to incorporate pre-trained models or existing data).   

In this paper, we develop \textit{learning to defer to a population}: an L2D system that can accurately defer to humans whose predictions were not observed during training.  To achieve this, our L2D system is not customized to individuals but rather a \emph{population}.  At test time, we assume that an expert will be drawn from this population, and the L2D system needs to make appropriate deferral decisions despite some uncertainty in how this specific expert will behave.  We develop surrogate loss functions for this setting, assuming we have access to similarly-sampled experts to train from, and show that they are consistent.  

For a general implementation, we propose \textit{meta-learning} \citep{Schmidhuber1987, thrun1998lifelong} on a small context set that is representative of the expert's abilities.  We consider two approaches: one is to fine-tune a model that represents a typical expert, and the other is to encode the context set with a deep-sets architecture \citep{zaheer2017deep}.  We perform experiments on image recognition, traffic sign detection, and skin lesion diagnosis tasks, showing that our models are able to perform well even as expert variability increases.

\section{BACKGROUND}
We begin by reviewing the L2D framework for both single \citep{pmlr-v119-mozannar20b} and multiple \citep{verma2023learning} experts.  For the whole of this paper, we consider only multi-class classification, but all methods can be straightforwardly extended to other data types, such as real-valued regression \citep{zaoui2020regression}.

\subsection{Single-Expert Setting}
\paragraph{Data \& Models} We first define the data for multiclass L2D with one expert.  Let $\mathcal{X}$ denote the feature space, and let $\mathcal{Y}$ denote the label space, a categorical encoding of multiple ($K$) classes.  $\rvx_n \in \mathcal{X}$ denotes a feature vector, and $\ry_n \in \mathcal{Y}$ denotes the associated class defined by $\mathcal{Y}$ (1 of $K$).  In order to model the expert's abilities, L2D assumes that we have access to (human) expert demonstrations.  Denote the expert's prediction space as $\mathcal{M}$ = $\mathcal{Y}$, and let the demonstrations be denoted $\rsm_n \in \mathcal{M}$ for the associated features $\rvx_n$.  The $N$-element training sample is then $\mathcal{D} = \{\vx_n, y_n, m_n\}_{n=1}^{N}$.  As for model specification, the L2D framework requires two sub-models: a classifier and a rejector \citep{cortes2016learning,cortes2016boosting}.  Denote the \textit{classifier} as $h: \mathcal{X} \rightarrow \mathcal{Y}$ and the \textit{rejector} as $r: \mathcal{X} \rightarrow \{0,1\}$.  The rejector can be interpreted as a meta-classifier, determining which inputs are appropriate to pass to $h(\rvx)$.  When $r(\rvx)=0$, the classifier makes the decision, and when $r(\rvx)=1$, the classifier abstains and defers the decision to the human.

\paragraph{Learning} The learning problem requires fitting both the rejector and classifier.  When the classifier makes the prediction, then the system incurs a loss of zero (correct) or one (incorrect).  When the human makes the prediction (i.e.~$r(\vx)=1$), then the human also incurs the same 0-1 loss.  Using the rejector to combine these losses, we have the overall classifier-rejector loss: 
\begin{equation}\label{eq:0-1}\begin{split}
 &L_{0-1}(h, r) = \\ & \ \ \mathbb{E}_{\rvx, \ry, \rsm}\left[(1-r(\rvx)) \  \mathbb{I}[h(\rvx) \ne \ry] \ + \ r(\rvx) \ \mathbb{I}[\rsm \ne \ry] \right]
\end{split}
\end{equation} where $\mathbb{I}$ denotes an indicator function that checks if the prediction and label are equal.  Upon minimization, the resulting Bayes optimal classifier and rejector satisfy: 
\begin{equation}\begin{split}\label{eq:Bayes_optimal_rej_clf_0-1_loss}
    h^{*}(\vx) &= \argmax_{y \in \mathcal{Y}} \ \mathbb{P}(\ry = y | \vx), \\
    r^{*}(\vx) &= \mathbb{I}\left[\mathbb{P}(\rsm = \ry | \vx) \ge \max_{y \in \mathcal{Y}} \mathbb{P}(\ry = y | \vx) \right],
\end{split}
\end{equation} where $\mathbb{P}(\ry | \vx)$ is the probability of the label under the data generating process, and $\mathbb{P}(\rsm = \ry | \vx)$ is the probability that the expert is correct.  The expert likely will have knowledge not available to the classifier, and this assumption allows the expert to possibly outperform the Bayes optimal classifier.

\paragraph{Softmax Surrogate} A consistent surrogate loss for the above $L_{0-1}$ loss can be derived following \citet{pmlr-v119-mozannar20b}'s formulation. First let the classifier and rejector be unified via an augmented label space that includes the rejection option: $\mathcal{Y}^{\bot} = \mathcal{Y} \cup \{\bot\}$, where $\bot$ denotes the rejection option.  Secondly, let $g_{k}:\mathcal{X} \mapsto \mathbb{R}$ for $k \in [1, K]$ where $k$ denotes the class index, and let $g_{\bot}:\mathcal{X} \mapsto \mathbb{R}$ denote the rejection ($\bot$) option.  These $K+1$ functions can be combined using a loss that resembles the cross-entropy loss for a softmax parameterization:
\begin{equation}\label{eq:sm_loss}\begin{split}
     \rphi_{\text{SM}}(g_{1}&,\ldots, g_{K}, g_{\bot}; \vx, y, m) =  \\ & -\log \left( \frac{\exp\{g_{y}(\vx)\}}{\sum_{y' \in \mathcal{Y}^{\bot}} \exp\{g_{y'}(\vx)\} }\right) \\ & - \mathbb{I}[m = y]  \ \log \left( \frac{\exp\{g_{\bot}(\vx)\}}{\sum_{y' \in \mathcal{Y}^{\bot}} \exp\{g_{y'}(\vx)\} }\right).
\end{split}
\end{equation} The intuition is that the first term maximizes the function $g_{k}$ associated with the true label.  The second term then maximizes the rejection function $g_{\bot}$ but only if the expert's prediction is correct.  At test time, the classifier is obtained by taking the maximum over $k \in [1, K]$: $\hat{y} = h(\vx) = \argmax_{k \in [1, K]} g_{k}(\vx)$.  The rejection function is similarly formulated as $r(\vx) = \mathbb{I}[ g_{\bot}(\vx) \ge \max_{k} g_{k}(\vx) ]$.  The minimizers $g^{*}_{1},\ldots, g^{*}_{K}, g^{*}_{\bot}$ of $\rphi_{\text{SM}}$ also uniquely minimize $L_{0-1}(h, r) $, the $0-1$ loss from Equation \ref{eq:0-1} \citep{pmlr-v119-mozannar20b}.  \citet{verma2022calibrated} showed that using a one-vs-all parameterization and an analogous loss function is also a consistent surrogate and better estimates the expert's probability of being correct.  %

\subsection{Multi-Expert Setting}

\paragraph{Data \& Model} Now let there be $J$ experts, each having a prediction space denoted $\mathcal{M}_{j}$ = $\mathcal{Y}$ $ \ \forall j$.  Analogously to above, the expert demonstrations are denoted $\rsm_{n,j} \in \mathcal{M}_{j}$ for the associated features $\rvx_n$.  The combined $N$-element training sample then includes the feature vector, label, and all expert predictions: $\mathcal{D} = \{\vx_n, y_n, m_{n,1}, \ldots, m_{n,J}\}_{n=1}^{N}$.  In single-expert L2D, the rejector makes a binary decision---to defer or not---but in multi-expert L2D, the rejector also chooses \emph{to which} expert to defer.  In turn, define the multi-expert rejector as $r: \mathcal{X} \rightarrow \{0,1,\ldots, J\}$.  When $r(\rvx)=0$, the classifier makes the decision, and when $r(\rvx)=j$, the classifier abstains and the $j$th expert makes the prediction.  The classifier sub-model ($h$) is identical to the single-expert case.

\paragraph{Learning} Again the learning objective is to apply the $0-1$ loss to each decision maker:
\begin{equation*}\begin{split}
 &L_{0-1}(h, r) =  \mathbb{E}_{\rvx, \ry, \{\rsm_j\}_{j=1}^{J}}\Bigg[\mathbb{I}[r(\rvx) = 0] \  \mathbb{I}[h(\rvx) \ne \ry] \\ & \ \ \ \ \ \ \ \ \ \ \ \ \ \ \ \ \ \ \ \ \ \ \ \ \ \   + \ \sum_{j=1}^{J} \mathbb{I}[r(\rvx) = j] \ \mathbb{I}[\rsm_{j} \ne \ry] \Bigg]
\end{split}
\end{equation*}
The Bayes optimal classifier is the same as in the single-expert setting (Equation \ref{eq:Bayes_optimal_rej_clf_0-1_loss}). The optimal rejector is: 
\begin{equation*}\begin{split}
    r^{*}(\vx) &= \begin{cases} &0 \text{ if } \mathbb{P}(\ry = h^{*}(\vx) | \vx) > \mathbb{P}(\rsm_{j'} = \ry | \vx) \ \  \forall j'\\ &\argmax_{j \in [1, J]} \mathbb{P}(\rsm_{j} = \ry | \vx) \ \  \text{ otherwise},
\end{cases}
\end{split}
\end{equation*} where $\mathbb{P}(\ry | \vx)$ is again the probability of the label under the data generating process and $\mathbb{P}(\rsm_{j} = \ry | \vx)$ is the probability that the $j$th expert is correct.

\paragraph{Softmax Surrogate Loss} 
Lastly we define the multi-expert analog of the softmax-based surrogate loss \citep{verma2023learning}.  Define the augmented label space as $\mathcal{Y}^{\bot} = \mathcal{Y} \cup \{\bot_{1}, \ldots, \bot_{J} \}$ where $\bot_{j}$ denotes the decision to defer to the $j$th expert.  Again let the classifier be composed of $K$ functions: $g_{k}:\mathcal{X} \mapsto \mathbb{R}$ for $k \in [1, K]$ where $k$ denotes the class index.  The rejector can be implemented with $J$ functions: $g_{\bot, j}:\mathcal{X} \mapsto \mathbb{R}$ for $j \in [1, J]$ where $j$ is the expert index.  These $K+J$ functions can then be combined using a softmax-parameterized surrogate loss:
\begin{equation*}\begin{split}
     &\rphi_{\text{SM}}^{J}\left(g_{1},\ldots, g_{K}, g_{\bot, 1}, \ldots, g_{\bot, J}; \vx, y, m_{1}, \ldots, m_{J}\right) =  \\ & -\log \left( \frac{\exp\{g_{y}(\vx)\}}{\sum_{y' \in \mathcal{Y}^{\bot}} \exp\{g_{y'}(\vx)\} }\right) \\ & - \sum_{j=1}^{J} \mathbb{I}[m_{j} = y]  \ \log \left( \frac{\exp\{g_{\bot,j}(\vx)\}}{\sum_{y' \in \mathcal{Y}^{\bot}} \exp\{g_{y'}(\vx)\} }\right).
\end{split}
\end{equation*} The first term maximizes the function $g_{k}$ associated with the true label, and the second term maximizes the rejection function $g_{\bot, j}$ for every expert whose prediction is correct.  At test time, the classifier is obtained by taking the maximum over the first $K$ dimensions.  Deferral is determined according to $$r(\vx) = \begin{cases}
  &0 \ \ \text{ if } \ g_{h(\vx)} > g_{\bot,j'} \ \  \forall j' \in [1, J]\\
  &\argmax_{j \in [1, J]} g_{\bot,j}(\vx) \ \  \text{ otherwise}.
\end{cases}$$ 

\subsection{Meta-Learning}
\textit{Meta-learning} (or \textit{learning-to-learn}) \citep{Schmidhuber1987, thrun1998lifelong} is a framework that assumes there is a pool of multiple (possibly infinite) related tasks.  Since these tasks are assumed to share an underlying structure, one model is trained on all tasks so that information can be shared, i.e.~learning across learning problems.  Given a model $p(\mathfrak{D}_{t}; \rvtheta)$ where $\mathfrak{D}_{t}$ denotes the data for task $t$, meta-learning can be formulated as optimizing $ \vtheta^{*} \ = \ \argmax_{\rvtheta \in \Theta} \ \mathbb{E}_{\mathfrak{D}_{t}} \left[\log p\left(\mathfrak{D}_{t}; \rvtheta\right)\right],$ where the expectation is taken w.r.t.~$\mathbb{P}(\mathfrak{D}_{t})$, the generative process for all tasks.  Common approaches to meta-learning are based on metric learning \citep{vinyals2016matching}, meta-modeling \citep{santoro2016meta}, and meta-optimization \citep{ravi2017optimization}.  We will employ the latter two.  In both cases, we make the standard assumption that a \textit{context set} is available that describes the task at hand by way of a few exemplar data points: $\mathcal{D}_{t} = \{ (\vx_{b}, y_{b}) \}_{b=1}^{B}$.

\paragraph{Meta-Learning via Optimization} Optimization-based approaches take inspiration from \textit{fine-tuning}.  Given a pre-trained model, we could fine-tune it for a test-time task via gradient descent using $\mathcal{D}_{t}$.  Yet fine-tuning on a small context set makes it hard to balance the trade-off between adapting to the new task vs leveraging the knowledge from pre-training \citep{ravi2017optimization}.  In turn, much of the work on optimization-based meta-learning has proposed learning rules that better manage this trade-off.  Examples include LSTM-inspired gating of the gradient update \citep{ravi2017optimization}, simulating fine-tuning during training \citep{finn2017model}, and using models to define black-box parameter updates \citep{andrychowicz2016learning}.  See \citet{hospedales2021meta} for a survey.  We consider traditional fine-tuning and \textit{model-agnostic meta-learning} \citep{finn2017model} for our implementations but other approaches are applicable.

\paragraph{Neural Processes} Our model-based approach will employ an architecture similar to the \textit{conditional neural process} (CNP) \citep{garnelo2018conditional}.  The CNP parameterizes a predictive model $p(\ry | \rx, \mathcal{D}_{t})$ where $\ry$ is a label, $\rx$ a feature vector, and $\mathcal{D}_{t}$ the context set.  The CNP supports fast adaptation to $\mathcal{D}_{t}$ by passing it through a permutation-invariant deep set encoder \citep{zaheer2017deep}.  The representation produced by this encoder is passed along with the feature vector $\rvx$ into a decoder that parameterizes the predictive distribution for $\ry$.  Neural processes have been extended in a variety of ways \citep{gordon2019convolutional, wang2022moe, foong2020meta, kawano2021group, tailor2023exploiting}; one that we will also consider is an attention-based variant \citep{kim2019attentive}.  The key feature of the attentive neural process is that it applies cross-attention between $\rvx$ and $\mathcal{D}_{t}$, allowing the current feature vector to up-weight particular points in the context set that seem most relevant for making the current prediction.

\section{LEARNING TO DEFER TO A POPULATION}\label{sec:l2d_pop}
Following the HI paradigm, we wish to have a L2D system that can call upon a human expert for help in difficult cases.  Yet all existing L2D systems assume that the expert who is available at test time is the same as the one who provided training data \citep{leitao2022human}.  Many real-world settings do not support such an assumption.  Consider deploying an L2D system for radiology: a technician performs the medical imaging, and given these images, the L2D system can either make a diagnosis itself or defer the decision to the on-call radiologist.  For an existing L2D system to be successful, demonstrations from the available radiologist must have been included in the training data.  However, there are many plausible scenarios in which test-time decision making will differ from the training conditions.  For example, maybe a radiologist from a neighboring hospital is filling in due to a staff shortage, perhaps a new radiologist has recently joined the staff, or maybe even a radiologist who was observed during training has started to suffer some form of mental decline (e.g.~from illness).  

While generalizing to unseen experts may seem daunting, progress can be made by assuming all possibly-available experts have commonalities in their decision making.  Returning to the healthcare example, all radiologists who might work at the hospital have presumably received similar training and certifications.  This shared knowledge results in a coherent statistical signal that admits learning from the underlying static \emph{population} of experts.  We next describe how to adapt the L2D framework to be robust to a random expert at test time, so long as that expert is drawn from the same population as those seen during training.

\subsection{Theoretical Formulation}
\paragraph{Generative Process for Experts} We describe the above motivating setting with the following hierarchical generative process.  Let $\mathfrak{E}$ denote a random variable that represents a particular \emph{expert}.  The generative process for this expert's prediction $\rsm$, for an $(\rvx, \ry)$ pair, is: 
\begin{equation}\label{eq:gen_proc}
    \mathfrak{E} \sim \mathbb{P}(\mathfrak{E}), \ \ \ \ \rsm \sim \mathbb{P}\left(\rsm | \rvx, \ry, \mathfrak{E}\right),
\end{equation} where $\mathbb{P}(\mathfrak{E})$ defines a \emph{population of experts} from which we can sample experts indefinitely and without repetition.  This assumption is directly motivated by wanting to generalize to never-before-seen experts.  The assumption that the expert's prediction is conditioned on the label is inherited from single-expert L2D, which also assumes $\rsm \sim \mathbb{P}\left(\rsm | \rvx, \ry\right)$ (see \citet{pmlr-v119-mozannar20b}'s Equation 1).  We term this variant of the L2D framework \textit{learning to defer to a population} (L2D-Pop).  

\paragraph{Models \& Learning}
The learning problem can be formulated similarly to single-expert L2D.  Again the model is comprised of a classifier, $h: \mathcal{X} \rightarrow \mathcal{Y}$, and a \textit{rejector}.  However, now the reject needs to take as input both the feature vector and some representation of the currently-available expert: $r: \mathcal{X} \times \mathfrak{E} \rightarrow \{0,1\}$.  Note this formulation's difference from multi-expert L2D, whose rejector has a $(J+1)$-dimensional range and scales linearly with the number of experts.  Applying the 0-1 loss to each decision maker, we have: \begin{equation}\label{eq:0-1_expert_cond}\begin{split}
 L_{0-1}(h, r) \ = \  &\mathbb{E}_{\rvx, \ry, \rsm, \mathfrak{E} }\Big[(1-r(\rvx, \mathfrak{E})) \  \mathbb{I}[h(\rvx) \ne \ry] \\ &  \ \ \ \ \ \ \ \ \ \ \ \ \ \ \ \ \ \   + \ r(\rvx, \mathfrak{E}) \ \mathbb{I}[\rsm \ne \ry] \Big].
\end{split}
\end{equation} The difference from Equation \ref{eq:0-1} is now the expectation and rejector both include the expert variable $\mathfrak{E}$.

\paragraph{Bayes Solutions}  We derive the Bayes optimal classifier and rejector for Equation \ref{eq:0-1_expert_cond} in \cref{app:theory}. The optimal classifier, unsurprisingly, is the same as in single- and multi-expert L2D (Equation \ref{eq:Bayes_optimal_rej_clf_0-1_loss}).  The difference resides in the optimal rejector, as it now is a function of $\mathbb{P}\left( \rsm = \ry | \rvx, \mathfrak{E} \right)$, the probability that a particular expert $\mathfrak{E}$ will correctly predict $\ry$: \begin{equation}\begin{split}\label{eq:Bayes_optimal_crowd_cond}
    r^{*}(\vx, \mathfrak{E}) &= \mathbb{I}\left[\mathbb{P}\left( \rsm = \ry | \vx, \mathfrak{E} \right) \ge \max_{y \in \mathcal{Y}} \mathbb{P}(\ry = y | \vx) \right].
\end{split}
\end{equation} 

\subsection{Consistent Surrogate Losses}
We now discuss how to implement L2D-Pop, describing both the required data and the form of the consistent surrogate losses.

\paragraph{Data}  Assume we observe a training set of $N$ data points.  Each feature-label pair is associated with $E_{n}$ expert demonstrations and expert representations, which we denote as $\rvpsi^{\mathfrak{E}}$: $$\mathcal{D} = \left\{ \vx_{n}, y_{n}, \left\{ m_{n, e}, \vpsi^{\mathfrak{E}}_{e} \right\}_{e=1}^{E_{n}}\right\}_{n=1}^{N}.$$  The subscript $n$ in $E_{n}$ means that the number of experts who have provided demonstrations can change for every data point.  Moreover, two data points can have demonstrations provided from non-overlapping sets of experts.  Having a variable number of experts in this way is not permitted by existing, provably-consistent L2D systems, but non-theoretically-grounded approaches have been proposed for this problem \citep{hemmer2023learning}.    %

\paragraph{Surrogate Losses}  We show the consistency of both the softmax- and OvA-based surrogates for L2D-Pop in \cref{app:theory}.  The proofs follow a similar recipe to the single- and multi-expert L2D setting, by again defining the augmented label space $\mathcal{Y}^{\bot}$ and using a reduction to cost-sensitive learning.  Again we define $K+1$ functions: $g_{k}:\mathcal{X} \mapsto \mathbb{R}$ for $k \in [1, K]$, where $k$ denotes the class index, and $g_{\bot}:\mathcal{X} \mapsto \mathbb{R}$ denotes the rejection ($\bot$) score.  Combining these functions via the softmax function, the consistent surrogate has the form:
\begin{equation}\label{eq:sm_loss_expert_cond}
\begin{split}
     &\rphi_{\text{SM-Pop}}\left(g_{1},\ldots, g_{K}, g_{\bot}; \vx, y,  \left\{ m_{e}, \vpsi^{\mathfrak{E}}_{e} \right\}_{e=1}^{E} \right) =  \\ & \sum_{e=1}^{E} -\log \left( \frac{\exp\{g_{y}(\vx)\}}{\mathcal{Z}(\vx, \vpsi^{\mathfrak{E}}_{e})} \right)  \\ & -  \  \mathbb{I}[m_{e} = y] \ \log \left( \frac{\exp\{g_{\bot}(\vx, \vpsi^{\mathfrak{E}}_{e})\}}{\mathcal{Z}(\vx, \vpsi^{\mathfrak{E}}_{e}) }\right), \\ & \text{where } \\ & \mathcal{Z}(\vx, \vpsi^{\mathfrak{E}}_{e}) = \exp\{g_{\bot}(\vx, \vpsi^{\mathfrak{E}}_{e})\} + \sum_{y' \in \mathcal{Y}} \exp\{g_{y'}(\vx)\}. 
\end{split}
\end{equation} The rejector, crucially, is a function of both the features and the expert representation $\vpsi^{\mathfrak{E}}_{e}$.  See \cref{app:theory} for the OvA surrogate.  

The simplest way to implement the expert representation $\vpsi^{\mathfrak{E}}_{e}$ is to use tabular meta-data describing this expert.  In the radiology example, $\vpsi^{\mathfrak{E}}_{e}$ could be a vector containing information such as the number of years the doctor has been practicing, what certifications and training they have received, the quality of the diagnoses they have provided in the past, etc.  In Section \ref{sec:meta-L2D}, we describe a general meta-learning approach that encodes the expert's representation directly from their past decision making.  But first we describe how one could apply the single-expert framework to L2D-Pop.

\subsection{Applying Single-Expert L2D to L2D-Pop}\label{sec:single-to-pop}
While multi-expert L2D cannot be applied to L2D-Pop, we can consider the natural baseline of applying single-expert L2D to model the \emph{average expert} defined by the population.  Instead of having the rejector adapt to a particular expert $\mathfrak{E}$, it can model the population's \emph{marginal} probability of correctness: $$\mathbb{P}_{\mathfrak{E}}\left( \rsm = \ry | \rvx \right) = \int_{\mathfrak{E}} \mathbb{P}\left( \rsm = \ry | \rvx, \mathfrak{E} \right) \  \mathbb{P}\left( \mathfrak{E}\right) d \mathfrak{E},$$ where $\mathbb{P}\left( \rsm = \ry | \rvx, \mathfrak{E} \right)$ is the same quantity from Equation \ref{eq:Bayes_optimal_crowd_cond}, and the expert is marginalized away.  We use the subscript $\mathfrak{E}$ in $\mathbb{P}_{\mathfrak{E}}$ to make clear the probability is for a given population and to distinguish this quantity from the single-expert setting ($\mathbb{P}(\rsm = \ry | \vx)$). 

This modification of single-expert L2D can be done straightforwardly by including the distribution over experts in Equation \ref{eq:0-1}: 
\begin{equation*}\begin{split}
 &L_{0-1}(h, r) = \\ & \mathbb{E}_{\rvx, \ry, \rsm, \mathfrak{E} }\left[(1-r(\rvx )) \  \mathbb{I}[h(\rvx) \ne \ry] \ + \ r(\rvx  ) \ \mathbb{I}[\rsm \ne \ry] \right].
\end{split}
\end{equation*}  The Bayes optimal classifier again remains the same (Equation \ref{eq:Bayes_optimal_rej_clf_0-1_loss}), but the optimal rejector now includes the marginal correctness term from above:
\begin{equation}\begin{split}\label{eq:Bayes_optimal_crowd_single}
    r^{*}(\vx) &= \mathbb{I}\left[\mathbb{P}_{\mathfrak{E}}\left( \rsm = \ry | \vx \right) \ge \max_{y \in \mathcal{Y}} \mathbb{P}(\ry = y | \vx) \right].
\end{split}
\end{equation}  Both the single-expert softmax and OvA surrogate losses can be easily adapted to this setting.  For instance, the softmax surrogate from Equation \ref{eq:sm_loss} can be re-formulated for L2D-Pop as:
\begin{equation*}
\begin{split}
     &\rphi_{\text{SM-Pop-Avg}}\left(g_{1},\ldots, g_{K}, g_{\bot}; \vx, y,  \left\{ m_{e} \right\}_{e=1}^{E} \right) =  \\ & -\log \left( \frac{\exp\{g_{y}(\vx)\}}{\sum_{y' \in \mathcal{Y}^{\bot}} \exp\{g_{y'}(\vx)\} }\right) \\ & - \left(\frac{1}{E}\sum_{e=1}^{E} \mathbb{I}[m_{e} = y]\right) \ \log \left( \frac{\exp\{g_{\bot}(\vx)\}}{\sum_{y' \in \mathcal{Y}^{\bot}} \exp\{g_{y'}(\vx)\} }\right).
\end{split}
\end{equation*} where $\frac{1}{E}\sum_{e=1}^{E} \mathbb{I}[m_{e} = y]$, the fraction of the population that correctly predicted this point, is an empirical estimate of $\mathbb{P}_{\mathfrak{E}}\left( \rsm = \ry | \rvx \right)$.  This loss is similar to Equation \ref{eq:sm_loss_expert_cond}, but the rejector term $g_{\bot}$ is no longer a function of the expert.  Thus the sum over experts `pushes through' to just the indicator term $\mathbb{I}[m_{e} = y]$.

\begin{figure*}
    \centering
    \includegraphics[width=.8\linewidth]{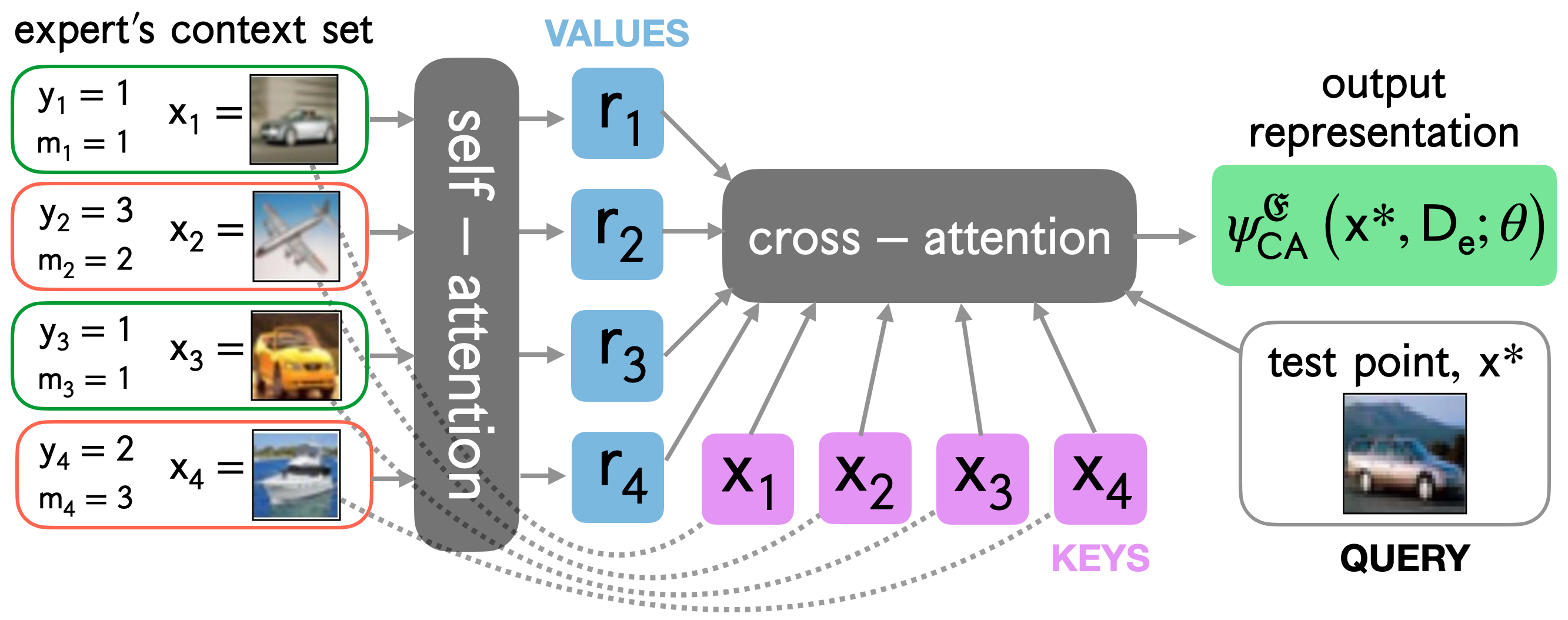}
    \caption{\textit{Attentive Encoder of Expert's Context Set.} The above diagram shows how an expert's context set is summarized into a representation.  The cross-attention mechanism allows points in the context set to be emphasized if they are similar to the current query point.  In the example above, images of cars would be emphasized to determine if this expert performs well at classifying cars.}
    \label{fig:attention_diagram}
    \vspace*{-1em}
\end{figure*}

\section{META-LEARNING TO DEFER}\label{sec:meta-L2D}
While we have given a complete recipe for L2D-Pop, the above implementation relies upon having an effective way to summarize the expert via the feature vector $\vpsi^{\mathfrak{E}}$.  However, specifying these features will greatly depend on the application and having domain knowledge.  For cases without strong prior knowledge, we now describe a general meta-learning approach that allows the rejector to leverage whole (but likely small) data sets that are representative of the current expert's decision making. 

\paragraph{Context Set}  Instead of using handcrafted features, we now assume that the model has access to a small but representative set of demonstrations for any given expert.  Denote this \textit{context set} for the $e$-th expert as:  $$\mathcal{D}_{e} = \left\{\rd_{e, b}\right\}_{b=1}^{B} = \left\{ (\rvx_{e,b}, \ry_{e, b}, \rsm_{e, b}) \right\}_{b=1}^{B}$$ where $B$ is the size of the set.  Ideally these demonstrations should have been collected recently.  This set should also be as large as possible without burdening the expert's time (what constitutes a `burden' will depend on the effort it takes the expert to produce each prediction).  For the radiology example, the context set could be obtained by having the doctor perform a few `warm-up' diagnoses (on historical data) before commencing her real shift.

\subsection{Meta-Optimization Approach}  
A straight-forward method for learning from the context set $\mathcal{D}_{e}$ is via meta-optimization.  In its most basic implementation, this approach takes the form of \textit{fine-tuning}.  First train the L2D-Pop system to model the marginal expert via the surrogate $\rphi_{\text{SM-Pop-Avg}}$ from Section \ref{sec:single-to-pop}.  Then at test time, when a new expert is available, perform gradient descent updates using $\mathcal{D}_{e}$.  This will adapt the L2D-Pop system to move away from modeling the marginal expert and (hopefully) towards modeling the newly available one.  

One can also employ \textit{model-agnostic meta-learning} (MAML) \citep{finn2017model} for L2D-Pop.  MAML aims to make the model amenable to test-time fine-tuning by simulating fine-tuning during training.  
Ultimately, we found that using MAML made it hard to balance the performance of the classifier and rejector; see \cref{app:maml} for further details and the results.  Traditional fine-tuning was more stable, and for this reason, we report its performance in the experiments as representative of the meta-optimization approach.   

\subsection{Model-Based Approach} 
We now describe a model-based approach to meta-learning for L2D-Pop.  Instead of using optimization to adapt the model, the entire context set will be used as input to a (data) set encoder.  This encoder then performs the adaptation via one forward pass.

\paragraph{Deep Sets Encoder}  Given the context set $\mathcal{D}_{e}$ describing the expert's decision making, we can encode it into a representation via a \textit{deep sets} architecture \citep{zaheer2017deep}.  While there are several choices available, we use the mean aggregation mechanism employed by \textit{neural processes} \citep{garnelo2018conditional}.  Let $\rvr = \rvgamma(\rd; \rvtheta)$ denote the output of a neural architecture (e.g.~multi-layer perceptron or ConvNet), with parameters $\rvtheta$, applied to a point in the expert's context set, $\rd$. Applying mean aggregation to the output of $\rvgamma$, the expert's representation is: $$ \vpsi^{\mathfrak{E}}\left( \mathcal{D}_{e}; \rvtheta \right) = \frac{1}{B} \sum_{b=1}^{B} \rvr_{e,b} = \frac{1}{B} \sum_{b=1}^{B} \rvgamma(\rd_{e,b}; \rvtheta),$$ where $\rvtheta$ needs to be fit along with the other parameters of the L2D system.  The deferral function is then composed with the output of the set encoder: $g_{\bot}\left(\vx, \vpsi^{\mathfrak{E}}\left( \mathcal{D}_{e}; \rvtheta \right)\right)$.

\paragraph{Cross-Attention Mechanism} One drawback of the mean pooling mechanism is that all points are weighted equally.  However, even if an expert is an overall poor decision maker, if they make good predictions on points similar to the current test point, then this could be reason enough to defer to them.  Following the \textit{attentive} neural process \citep{kim2019attentive}, we apply cross-attention between $\mathcal{D}_{e}$ and a test-point $\rvx$.  Figure \ref{fig:attention_diagram} shows a diagram of the computation, which also includes an optional self-attention mechanism to generate richer representations of $\mathcal{D}_{e}$.  In the figure, the test point $\rvx^{*}$ is an image of a car, and thus cross-attention should up-weight the first and third points in the context set, as they also contain cars.  We denote the output of the cross-attention mechanism as $\vpsi_{\text{CA}}^{\mathfrak{E}}\left(\rvx,  \mathcal{D}_{e}; \rvtheta \right)$, and in turn, the deferral function is $g_{\bot}\left(\rvx, \vpsi_{\text{CA}}^{\mathfrak{E}}\left(\rvx,  \mathcal{D}_{e}; \rvtheta \right)\right)$, where $\rvx$ is an input to both the deferral function and the expert representation learner.    

\paragraph{Missing Demonstrations at Test Time} One may worry that, at test time, an expert can appear who does not have a corresponding context set.  Such a situation is easy to handle with meta-optimization since the model can simply remain fixed to model the marginal expert (which is a good inductive bias).  On the other hand, there is no guarantee for how the set encoder will behave if given the empty set as input.  If this is a worry, one could train a second L2D system that models the marginal expert (from Section \ref{sec:single-to-pop}) and use it instead for expert's with empty context sets.  Yet, in \cref{app:missing_context} we show that our neural process does not defer when the context set is missing: coverage is nearly 99\%. This is an appropriate behavior since, if the model does not have any information about the available expert, then not deferring is a safe decision.  One could also try to use the imputation method of \citet{hemmer2023learning} to fill in the missing values.

\begin{figure*}
    \centering
    \includegraphics[width=.9\linewidth]{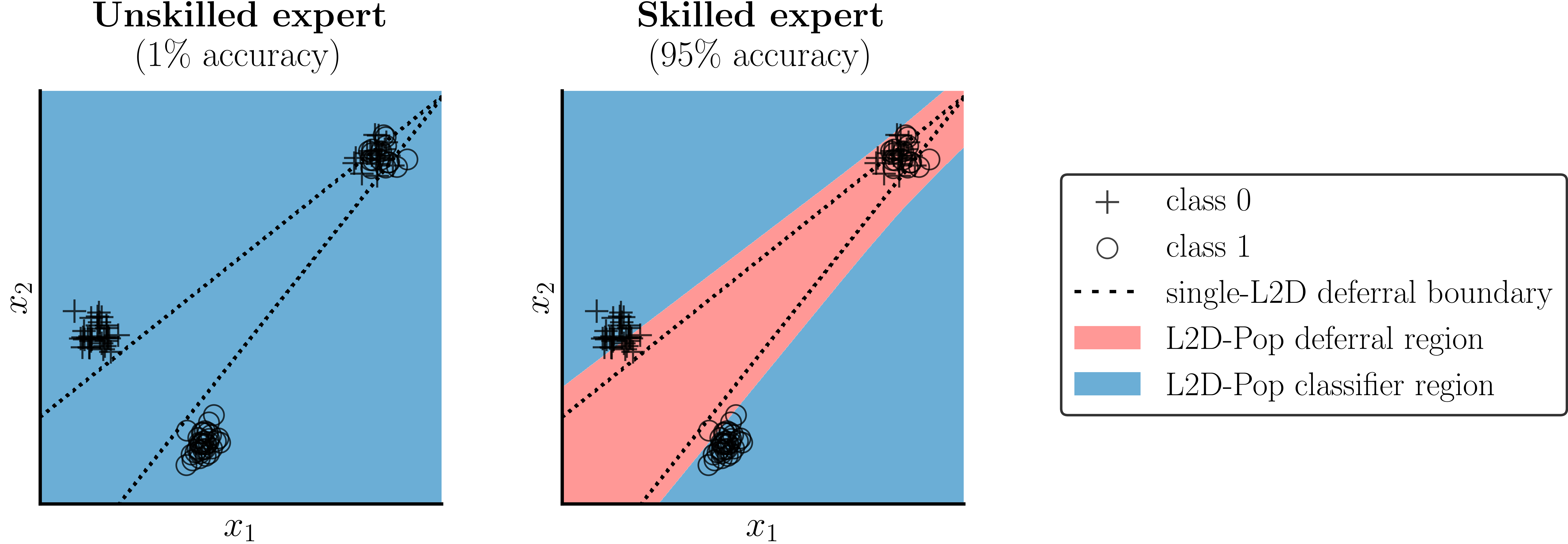}
    \caption{\textit{Synthetic 2D Data.} We simulate three clusters, two having class purity and a third having a mixture of two classes.  Furthermore, we simulate three experts and show  the model's decision regions for the worst ($1\%$) and best ($95\%$).  
    The dashed line is where single-expert L2D defers; it is constant across experts.  The red region is where L2D-Pop defers; it successfully adapts to the expert by never deferring the former case and deferring the whole of the difficult cluster in the latter case.}
    \label{fig:simulation}
\end{figure*}

\section{RELATED WORK}
Classifiers with the ability to reject or abstain have long been studied \citep{Chow1957AnOC}, with \citet{madras2018predict} developing the modern formulation that we consider.  The two primary approaches to making the rejection decision have been confidence-based \citep{bartlett2008class, yuan2010class, jiang2018trust, grandvalet2009svm, Ramaswamy2018ConsistentAF, Ni2019OnTC} and model-based \citep{cortes2016learning,cortes2016boosting}.  The theoretical properties of the classifier-rejector approach have been well-studied for binary \citep{cortes2016learning, cortes2016boosting} and multi-class classification \citep{Ni2019OnTC, pmlr-v139-charoenphakdee21a, pmlr-v119-mozannar20b, yuzhou2022generalizing}.  There are also various L2D relatives that do not come with consistency guarantees \citep{raghu2019algorithmic, Wilder2020LearningTC,  pradier2021preferential, okati2021differentiable, liu2022incorporating}.  Several limitations of the current L2D algorithms have also been studied, including mis-calibration \citep{verma2022calibrated, cao2023in}, under-fitting \citep{narasimhan2022posthoc}, realizable consistency \citep{mozannar2023exact}, sample complexity \citep{charusaie2022sample}, and data scarcity \citep{hemmer2023learning}.  As for L2D systems that support multiple experts, existing formulations \citep{keswani2021towards, hemmer2022forming, verma2023learning, mao2023twostage, mao2023principled} all consider a finite number of experts and assume they are seen during training.  Our meta-learning approach most resembles the work of \citet{hemmer2023learning}, as they try to cope with missing expert demonstrations via model-based imputation.  But they do not consider incorporating this model into a different or more general L2D framework, which is our primary contribution.

\section{EXPERIMENTS}\label{sec:exp}
\begin{figure*}[t!]
    \begin{subfigure}[t]{.32\linewidth}
        \centering
        \includegraphics{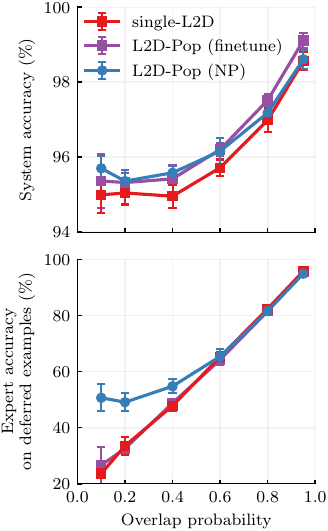}
        \caption{Traffic Signs}
    \end{subfigure}%
    \centering
    \begin{subfigure}[t]{.32\linewidth}
        \centering
        \includegraphics{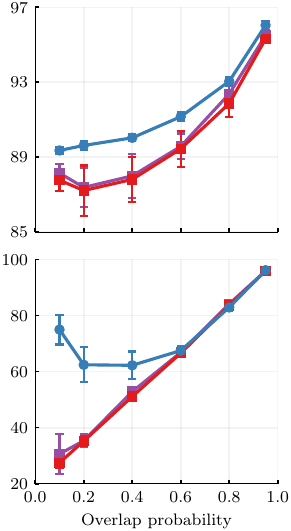}
        \caption{CIFAR-10}
    \end{subfigure}%
    \begin{subfigure}[t]{.32\linewidth}
        \centering
        \includegraphics{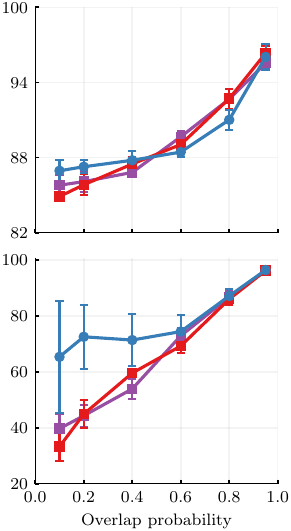}
        \caption{HAM10000}
    \end{subfigure}
    \caption{\textit{Varying Population Diversity on Image Classification Tasks.} L2D-Pop exploits experts' context sets to make better deferment decisions given by the increase in expert accuracy on deferred examples (bottom). This leads to a boost in overall system accuracy (top). The gap widens as the overlap in experts' abilities decreases.}
    \label{fig:image_classification}
    \vspace*{-1em}
\end{figure*}

We perform a range of experiments that isolate the effectiveness of L2D-Pop as the underlying population changes.  Our implementation is available at \url{https://github.com/dvtailor/meta-l2d}.  The primary baseline we considered is a single-expert L2D system (single-L2D) that models the population average, as described in Section \ref{sec:single-to-pop}.  This is an informative baseline since single-L2D and L2D-Pop will converge in performance as the population becomes concentrated around the mean expert.  We use the softmax-based surrogate loss for all results below.  Results for one-vs-all surrogates are included in \cref{app:exp_ova}.

\subsection{Synthetic 2D Data} %
We first perform a simulation to demonstrate the failure of modeling just the population average, as single-L2D does.  We simulate a binary classification task by drawing the features from one of three Gaussian distributions ($\vmu = \{[10, 10], [2,6], [6,2]\}$, $\Sigma = \text{diag}\{[0.2, 0.2]\}$), with the Gaussian located at $[10, 10]$ having a 50-50 mixture of classes and the other two clusters having $100\%$ class purity.  We construct three experts who make predictions across the feature space with a uniform correctness probability of $\{0.01, 0.80, 0.95\}$.  Here we consider just the model-based variant of L2D-Pop: the deferral function is parameterized by a neural process.  We use a linear model for the classifier, and the context embedding network and rejector network are parameterized by $3$-layer and $2$-layer MLPs respectively.  We sample $6000$ training examples, and context sets of size $B=60$ are sampled from the train set.  

Figure \ref{fig:simulation} shows the model's decision regions when the worst ($1\%)$ and best ($95\%$) experts are available.  Single-L2D is not adaptive and has the same decision region in both cases.  It fails by over-deferring in the former case, as the expert will do even worse than random chance on the third cluster.  Conversely, it under-defers in the latter case, as the model has only a random chance of being correct on the third cluster.  L2D-Pop successfully adapts to both settings, never deferring when the expert is poor and deferring the whole third cluster when the expert is good.

\begin{figure}[h]
    \centering
    \includegraphics{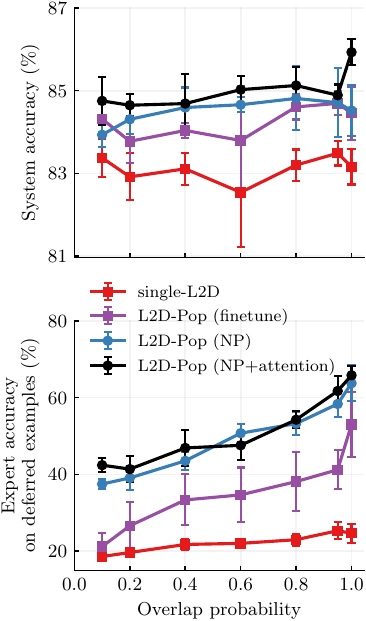}
    \caption{%
    L2D-Pop implemented with an attentive neural process (black) boosts performance when experts' abilities are specified by side-information (fine-grained labels) not provided in the context set.}
    \label{fig:cifar20}
    \vspace*{-1em}
\end{figure}

\subsection{Varying Population Diversity} \label{sec:exp_imageclass}
\paragraph{Data Sets}  We evaluate two variants of L2D-Pop, one implemented with meta-optimization (\emph{finetune}) and the other with model-based meta-learning (\emph{NP}).  For data sets, we use \texttt{CIFAR-10}, \texttt{Traffic Signs} \citep{houben2013detection}, and \texttt{HAM10000} \citep{tschandl2018ham10000} for skin lesion diagnosis. Our meta-learning models are compared against a L2D system that models the marginal expert (\emph{single-L2D}). 
 For \texttt{Traffic Signs}, we downsample the train set to 10,000 instances. During training, context sets are sampled from the train set, each containing 50 instances for \texttt{Traffic Signs} and \texttt{CIFAR-10}, and 140 for \texttt{HAM10000}. During evaluation, context sets are sampled from a 20\% split of the validation and test sets.  
 
 \paragraph{Models} We follow the approach of \citet{pmlr-v119-mozannar20b}, using a single base network. For L2D-Pop, the penultimate hidden activations of the base network are concatenated with the context embedding, and this is passed to a rejector network with a single output. 
The base network for \texttt{CIFAR-10} is a WideResNet with 28 layers and the base network for \texttt{HAM10000} is a ResNet-34.  
We warm-start these networks using model parameters from training the classifier only.  
The base network for \texttt{Traffic Signs} is a ResNet-20 (trained without warm-starting).
All networks are trained without data augmentation.  See \cref{app:pop_div} for further details on the training configuration. 

\paragraph{Expert Population} We construct synthetic experts by sampling without replacement classes for which the expert is an oracle. For non-oracle classes, the expert is correct with probability $p$ and otherwise predicts uniformly at random. $p$ is an \emph{overlap probability} which we increase from $0.1$ to $0.95$, toggling from specialized to identical experts, thereby representing the diversity of the expert population. We sample $10$ experts at train-time. For evaluation, we remove 5 of these and sample 5 new experts (simulating never-before-seen experts).  However, unlike during training where all ten experts are used, during evaluation only one expert is queried at a time (selected at random) for each new test point.  The number of oracle classes per expert is one for \texttt{CIFAR-10} and \texttt{HAM10000}, and five for \texttt{Traffic Signs} (due to it having many classes).

\paragraph{Results} Figure \ref{fig:image_classification} shows the combined accuracy of the classifier and expert versus the overlap probability $p$ (top) and the expert's accuracy on only deferred examples (bottom).  When the expert population is at its most diverse (far left), L2D-Pop is clearly superior at deferring, as shown by the expert accuracy on deferred examples, leading to an improved system accuracy over the single-expert baseline.  This improvement is more pronounced for the model-based (neural process) implementation.  The gap narrows as the overlap across experts increases, as is expected.  The neural process model also has faster adaptation as it does not need to run gradient updates at test-time; see \cref{tab:complexity} for a comparison of runtime.

\subsection{Ablation Study of Attention}
We study the effect of cross-attention in the neural process by taking \texttt{CIFAR-100} and merging the 100 classes into 20 superclasses of equal size. Unless otherwise stated, we follow the same setup as in Section \ref{sec:exp_imageclass}. We use context sets of size 100.
See \cref{app:exp_attention} for the details of the model architectures.  While the model predicts on the 20 superclasses, the 100 subclasses are used to construct experts with finer granularity. For each expert, we sample 4 superclasses uniformly at random without replacement. Then within these superclasses we pick 3 out of 5 subclasses at random. This gives a total of 12 subclasses for which the expert gives correct predictions. Outside of these subclasses, but within the superclasses, with overlap probability $p$ the expert is correct.  Otherwise the expert predicts uniformly at random amongst the superclasses. This hierarchical formulation allows us to experimentally verify that the model with cross-attention can better select experts by identifying if they are oracles for any subclasses within the superclasses.  Figure \ref{fig:cifar20} reports the results.  Cross-attention improves performance compared to the vanilla NP architecture, and especially compared to fine-tuning the marginal-expert model.

\section{CONCLUSIONS}
We proposed \textit{learning to defer to a population} (L2D-Pop), a generalization of learning to defer that allows for never-before-seen experts at test time.  This is achieved by training the model to generalize its deferral sub-component to all experts in a population.  We described two meta-learning implementations that adapt to any expert using a context set of demonstrations.  Our model is effective on data sets for traffic sign recognition and skin lesion diagnosis, especially as expert variability increases.  For future work, we plan to investigate alternative methods for meta-learning, such as metric-based approaches.  Moreover, the natural next step is to consider experts who change after training and therefore introduce distribution shift to the L2D problem.

\subsubsection*{Acknowledgements}
This publication is part of the project \textit{Continual Learning under Human Guidance} (VI.Veni.212.203), which is financed by the Dutch Research Council (NWO).  Putra Manggala was supported by the \textit{Hybrid Intelligence Center}, a $10$-year program funded by the Dutch Ministry of Education, Culture and Science through the Netherlands Organisation for Scientific Research.  Part of this work was carried out on the Dutch national e-infrastructure with the support of the SURF Cooperative. 

\bibliography{references}
\bibliographystyle{plainnat}

\newpage
\appendix
\onecolumn
\thispagestyle{empty}
{\hsize\textwidth
    \linewidth\hsize \toptitlebar {\centering
        {\Large\bfseries Learning to Defer to a Population: A Meta-Learning Approach\\Supplementary Material \par}}
    \bottomtitlebar}

\section{THEORETICAL RESULTS}\label{app:theory}
In this section, we provide proofs of the main results in the paper. The proofs follow from the results of \citet{verma2023learning}. We follow the notation from the paper. For simplicity, we assume all measure theoretic subtleties hold true. 

\subsection{Bayes Solution for L2D to a population}
The Bayes solution of L2D to a population follows directly from Proposition A.2 and Corollary A.3 of \citet{verma2023learning}. In particular, for an expert $\mathfrak{C} \sim \mathbb{P}\left(\mathfrak{C}\right)$ and $\rvx \sim \mathbb{P}(\rvx)$, their proposition can be extended to give the rejection rule $r^{*}\left(\rvx, \mathfrak{C}\right)$ as below:
\begin{align}
    \begin{split}
    r^{*}\left(\rvx, \mathfrak{C}\right) = 
        \begin{cases}
            1 \textrm{ if } \mathbb{E}_{\ry \vert \rvx}\left[\ell_{\text{clf}}\left(\hat{\ry}, \ry\right)\right] \geq \mathbb{E}_{\ry \vert \rvx} \mathbb{E}_{\rsm_{\mathfrak{C}} \vert \rvx, \ry, \mathfrak{C}}\left[\ell_{\text{exp}}\left(\rsm_{\mathfrak{C}}, \ry\right)\right] \forall \hat{y} \in \mathcal{Y}, \\
            0 \textrm{ otherwise },
        \end{cases}
    \end{split}
\end{align}
where $\hat{y}$ is the prediction of the classifier, $\ell_{\text{clf}}$ and $\ell_{\text{exp}}$ are respectively the loss functions for the classifier and the expert. In this paper, we consider the canonical $0-1$ loss, $\mathbb{I}\left[\hat{\ry} \neq \ry\right]$ (equivalently, $\mathbb{I}\left[\rsm_{\mathfrak{C}} \neq \ry\right]$), in which case the Bayes rejection rule follows immediately. 

\subsection{Consistency of $\rphi_{\text{SM-Pop}}$ (Equation \ref{eq:sm_loss_expert_cond})}
Define $-\log \left(\frac{\exp \{g_{y}\left(\vx\right)\}}{\mathcal{Z}\left(\vx, \mathbf{\psi}_{e}^{\mathfrak{C}}\right)}\right) = \zeta_{y}\left(\vx\right)$ and $-\log \left(\frac{\exp \{g_{\bot}\left(\vx, \mathbf{\psi}_{e}^{\mathfrak{C}}\right)\}}{\mathcal{Z}\left(\vx, \mathbf{\psi}_{e}^{\mathfrak{C}}\right)}\right)  = \zeta_{\bot, e}\left(\vx\right)$.
We consider the point-wise risk for $\rphi_{\text{SM-Pop}}$ written in terms of these terms as,
\begin{align*}
\begin{split}
    \mathcal{C}\left(\rphi_{\text{SM-Pop}}\right) &= \sum_{e=1}^{E}\left[\mathbb{E}_{\ry \vert \vx}\left[\zeta_{y}\left(\vx\right)\right] + \mathbb{E}_{\ry \vert \vx} \mathbb{E}_{\rsm \vert \vx, y, e} \left[\mathbb{I}\left[m_{e} = y\right]\cdot\zeta_{\bot, e}\left(\vx\right)\right]\right] \\
    &= \sum_{e=1}^{E}\left[\sum_{y \in \mathcal{Y}}\mathbb{P}\left(y\vert \vx\right)\zeta_{y}\left(\vx\right) + \sum_{y \in \mathcal{Y}}\mathbb{P}\left(y \vert \vx\right)\sum_{m_{e} \in \mathcal{Y}}\mathbb{P}\left(\rsm = m_{e} \vert \vx, y, e\right)\mathbb{I}\left[m_{e} = y\right]\cdot\zeta_{\bot, e}\left(\vx\right)\right] \\
    &= \sum_{e=1}^{E}\left[\sum_{y \in \mathcal{Y}}\mathbb{P}\left(y\vert \vx\right)\zeta_{y}\left(\vx\right) + \sum_{y \in \mathcal{Y}}\mathbb{P}\left(\ry = y \vert \vx\right)\mathbb{P}\left(\rsm = y \vert \vx, y, e\right)\cdot\zeta_{\bot, e}\left(\vx\right)\right] \\
    &= \sum_{e=1}^{E}\left[\sum_{y \in \mathcal{Y}}\mathbb{P}\left(y\vert \vx\right)\zeta_{y}\left(\vx\right) + \sum_{y \in \mathcal{Y}}\mathbb{P}\left(\rsm = y, \ry = y\vert \vx, e\right)\cdot\zeta_{\bot, e}\left(\vx\right)\right] \\
    &= \sum_{e=1}^{E}\left[\sum_{y \in \mathcal{Y}}\mathbb{P}\left(y\vert \vx\right)\zeta_{y}\left(\vx\right) + \mathbb{P}\left(\rsm = \ry \vert \vx, e\right)\cdot\zeta_{\bot, e}\left(\vx\right)\right].
\end{split}
\end{align*}
Considering that the above expression is a sum of $E$ convex terms, we can obtain the minimizer of the point-wise risk easily by differentiating the above expression w.r.t. $g_{y}\left(\vx\right)$ and $g_{\bot}\left(\vx, \psi_{e}^{\mathfrak{C}}\right)$. The crucial observation we make here is that $\boldsymbol{\psi}_{e}^{\mathfrak{C}}$ is a constant (the context associated with the expert if fixed). The said minimizers then satisfy the following conditions (for each $e$):
\begin{align*}
    \begin{split}
        &\frac{\partial \mathcal{C}\left[\rphi_{\text{SM-Pop}}\right]}{\partial g_{y}\left(\vx\right)} = 0 \implies \frac{\exp g_{y}\left(\vx\right)}{\mathcal{Z}\left(\vx, \boldsymbol{\psi}_{e}^{\mathfrak{C}}\right)} = \frac{\mathbb{P}\left(\ry = y \vert \rvx = \vx\right)}{1 - \mathbb{P}\left(\rsm = \ry \vert \vx, e\right)}, \textrm { and } \\
        &
        \frac{\partial \mathcal{C}\left[\rphi_{\text{SM-Pop}}\right]}{\partial g_{\bot}\left(\vx, \boldsymbol{\psi}_{e}^{\mathfrak{C}}\right)} = 0 \implies \frac{\exp g_{\bot}\left(\vx, \boldsymbol{\psi}_{e}^{\mathfrak{C}}\right)}{\mathcal{Z}\left(\vx, \boldsymbol{\psi}_{e}^{\mathfrak{C}}\right)} = \frac{\mathbb{P}\left(\rsm = \ry \vert \vx, e\right)}{1 - \mathbb{P}\left(\rsm=\ry \vert \vx, e\right)}.
    \end{split}
\end{align*}
Given how the rejection and prediction is set up in $\rphi_{\text{SM-Pop}}$ and these conditions, it follows that the minimizer of the point-wise risk adheres to the Bayes solution. Considering the hypothesis class of all functions, consistency follows.  

\subsection{One-vs-All (OvA) surrogate for L2D to a population}
In the main text, we considered softmax version of the surrogate loss. We can also extend the OvA version of the surrogate loss for L2D proposed by \citet{verma2022calibrated} for L2D to a population, as given below:
\begin{align*}
    \begin{split}
        &\rphi_{\text{OvA-Pop}}\left(g_1, \ldots, g_K; \vx, y, \{m_{e}, \boldsymbol{\psi}_{e}^{\mathfrak{C}}\}_{e=1}^{E}\right) = \\ & \sum_{e=1}^{E}\phi\left(g_{y}\left(\vx\right)\right) + \phi\left(-g_{\bot}\left(\vx, \boldsymbol{\psi}_{e}^{\mathfrak{C}}\right)\right) + \sum_{y' \in \mathcal{Y}/\{y\}}\phi\left(-g_{y}\left(\vx\right)\right) + \mathbb{I}\left[m_e = y\right]\left[\phi\left(g_{\bot}\left(\vx, \boldsymbol{\psi}_{e}^{\mathfrak{C}}\right)\right) - \phi\left(-g_{\bot}\left(\vx, \boldsymbol{\psi}_{e}^{\mathfrak{C}}\right)\right)\right],
    \end{split}
\end{align*}
where $\phi$ is some classification-calibrated \citep{bartlett2006convexity} binary surrogate loss function, e.g. the logistic loss. It is easier to establish the consistency of $\rphi_{\text{OvA-Pop}}$ following Theorem 1 in \citet{verma2022calibrated}. 

\section{EXPERIMENTAL DETAILS FOR IMAGE CLASSIFICATION TASKS} \label{app:image_class_exp_details}

\subsection{Varying Population Diversity} \label{app:pop_div}

The base network is trained using SGD with Nesterov momentum (default momentum parameter of 0.9). The other model components (final layer of the classifier, rejector and context embedding network) are trained using Adam. 
We use a cosine learning rate decay scheme, annealing the learning rate to $\frac{LR_{\text{cnn}}}{1000}$ and $\frac{LR}{1000}$ for the base network and remaining model components respectively, until $T_e\!-\!50$ epochs.
For the last $50$ epochs, the base learning rate is held constant at $\frac{LR_{\text{cnn}}}{1000}$ and $\frac{LR}{1000}$ for the respective model components.
We train without data augmentation nor do we use state-of-the-art network architectures as it is not our intention to obtain best classifier performance.

A checkpoint of the model is saved only if an improvement in the model performance is observed after every epoch. Nevertheless, training always runs for the full number of epochs stated but only the checkpointed model is evaluated on. 
We use the validation loss to score models except for HAM10000 where we use the system accuracy.
We use a batch size of 8 and 1 for the validation set and test set respectively -- for each minibatch, the expert is sampled anew and a new context set is drawn.
We also perform warmstarting for some of the experiments. 
The warmstart parameters are obtained by training the classifier only.
This is done in the same way as the L2D system except we train for a fixed number of epochs (100 for HAM10000 and 200 for the others) without intermediate model checkpointing and use a cosine learning rate decay scheme where the learning rate is annealed to 0.
A base learning rate of $10^{-2}$ is used for HAM10000 and the others use $10^{-1}$.
In the case of HAM10000, the warmstart parameters are obtained by fine-tuning from pretrained weights on ImageNet.
For fine-tuning the marginal single-expert L2D, we perform a grid search using the validation loss over the number of steps $[1,2,5,10,20]$ and step size $[10^{-3},10^{-2}]$ for HAM10000 and $[10^{-2},10^{-1}]$ for the others.
The fine-tuning is performed with vanilla gradient-descent where the base network parameters are frozen (i.e. only update the classifier and rejector final layer).
 
We now proceed to describe the context embedding network architecture.
First the base network is used to extract a feature vector for context inputs (corresponding to the penultimate layer activations).
We note that during training, we disable gradients backpropagating to the base network from the context embedding.
Next context labels and expert predictions are embedded using a linear layer with dimensionality $128$.
The above are then concatenated and passed to a MLP with width $128$ that outputs an embedding of $128$ dimensions.
This is repeated for all context points, the resulting embeddings for each context point are mean-pooled giving an aggregate embedding for the whole context set which acts as a proxy for the expert representation.
The rejector network is also given by a MLP with width $128$ that takes the extracted features for the test point (resulting from the base network) and the context embedding as input and outputs a single unit for the rejector logit.

The data sets are preprocessed as follows: for CIFAR-10, $10\%$ of the train set is used for the validation set. For Traffic Signs, the provided train set is downsampled to $10000$ examples and the provided test set is split $50{-}50$ into valid/test sets that we use. 
For HAM10000, the data set is prepared in the same way as \citep{verma2022calibrated} (60\% train, 20\% valid and 20\% test splits).
The experiments were run on an internal cluster of GPUs of the following type: Tesla V100 SXM2 16 GB.

\begin{table}[h!]
\begin{center}
\resizebox{\textwidth}{!}{%
\begin{tabular}{l*{11}{c}}
\toprule
Data Set & \makecell{Base\\Network} & Warm-Starting & \makecell{Batch\\Size} & $T_e$ & $K$ & $LR_{\text{cnn}}$ & $LR$ & $\delta$ & $B$ & $\ell_{\text{emb}}$ & $\ell_{\text{rej}}$\\
\midrule
Traffic Signs & ResNet-20 & \mycross & $64$ & $150$ & $43$ & $10^{-2}$ & $10^{-3}$ & $10^{-3}$ & $50$ & $5$ & $3$ \\
CIFAR-10 & WideResNet-28-2 & \mytick & $128$ & $100$ & $10$ & $10^{-2}$ & $10^{-3}$ & $5\!\times\! 10^{-4}$ & $50$ & $6$ & $4$\\
HAM10000 & ResNet-34 & \mytick & $128$ & $100$ & $7$ & $10^{-2}$ & $10^{-3}$ & $5\!\times\! 10^{-4}$ & $140$ & $6$ & $4$\\
CIFAR-20 & WideResNet-28-4 & \mytick & $128$ & $100$ & $20$ & $10^{-2}$ & $10^{-3}$ & $5\!\times\! 10^{-4}$ & $100$ & $6$ & $4$\\
\bottomrule
\end{tabular}}
\end{center}
\caption{Hyperparameters for image classification experiments: number of epochs $T_e$, number of classes $K$, initial learning rate of the CNN base network $LR_{\text{cnn}}$, initial learning rate of remaining model components $LR$, weight decay on CNN base network parameters $\delta$, context set size $B$, number of MLP layers for embedding network $\ell_{\text{emb}}$, number of MLP layers for rejector network $\ell_{\text{rej}}$.}
\label{tab:hyperparams}
\end{table}

\subsection{Ablation Study of Attention}\label{app:exp_attention}

Unless otherwise stated, the experimental setup follows \cref{app:pop_div} with specific hyperparameters given in \cref{tab:hyperparams}. We use data augmentation involving random horizontal flipping and random cropping. This is also used to obtain the warmstart parameters.
A separate base network is trained for the classifier and rejector.
The base network for the rejector is used to extract features for context inputs (and we also allow gradients to backpropogate in contrast to \cref{app:pop_div}).
We evaluate an additional meta-learning architecture where the context embedding network contains attention mechanisms (rejector network is unchanged).
This follows the design of the Attentive Neural Process (deterministic path) \citep{kim2019attentive} with a self-attention layer first applied over individual context embeddings followed by a cross-attention layer.
We use multi-head attention \citep{vaswani2017attention} with 8 heads throughout.
The number of parameters, training time and prediction time of this setting is reported in \cref{tab:complexity}.
This highlights a trade-off between training runtimes and speed of test-time adaptation in the fine-tuning and neural process approaches to L2D-Pop.

\begin{table}[h!]
\begin{center}
\begin{tabular}{l*{3}{c}}
\toprule
Method & \# Parameters & Training (s) & Prediction (s)\\
\midrule
single-L2D & 11698357 & 40.48 & 9.83\\
L2D-Pop (finetune) & 11698357 & 40.48 & 248.38\\
L2D-Pop (NP) & 11931317 & 148.90 & 14.57\\
L2D-Pop (NP w/ attention) & 12063413 & 163.33 & 15.11\\
\bottomrule
\end{tabular}
\end{center}
\caption{Number of parameters, training time and prediction time of L2D-Pop and the single-expert L2D baseline on the CIFAR-20 data set.
Training run time is measured for a single epoch (352 batches of size 128).
Prediction run time is measured over a full pass over the validation set (5000 examples) with batch size 8.
For \emph{L2D-Pop (finetune)}, this is only measured for a step size of $10^{-1}$ and 5 steps (time for grid search not included).
A NVIDIA GeForce RTX 3090 was used to obtain these runtimes.}
\label{tab:complexity}
\end{table}

\section{META-OPTIMIZATION USING MAML}
\label{app:maml}
\begin{figure*}[h]
    \centering
    \begin{subfigure}[h]{\linewidth}
        \centering
        \includegraphics{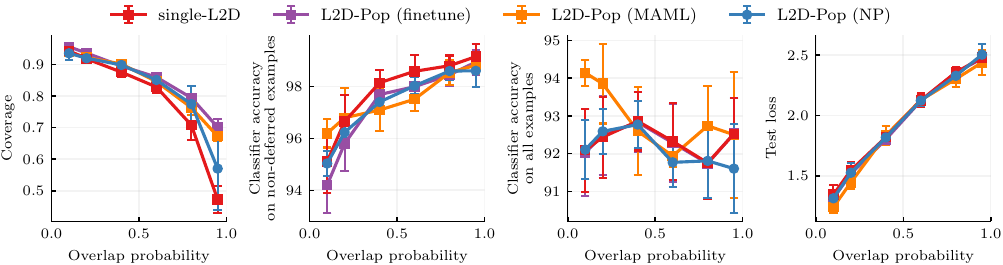}
        \caption{Traffic Signs}
    \end{subfigure}
    \begin{subfigure}[h]{\linewidth}
        \centering
        \includegraphics{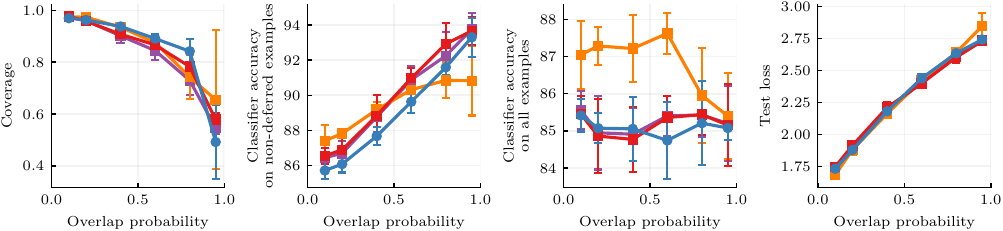}
        \caption{CIFAR-10}
    \end{subfigure}
    \begin{subfigure}[h]{\linewidth}
        \centering
        \includegraphics{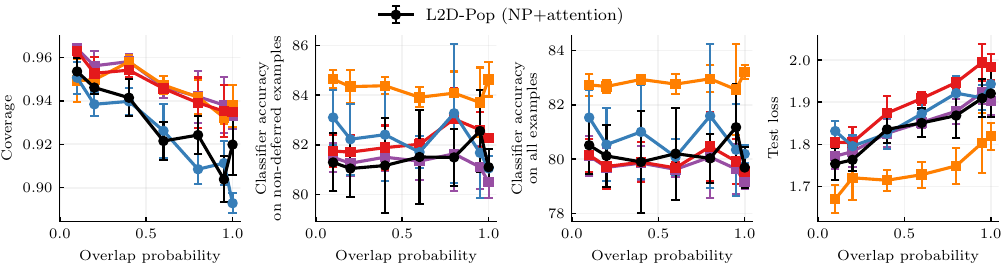}
        \caption{CIFAR-20}
    \end{subfigure}
    \caption{Additional metrics for MAML experiment.}
    \label{fig:image_classification_maml_extra}
\end{figure*}

We evaluate an additional meta-optimization approach to L2D-Pop using MAML (see \cref{alg:l2d_maml}).
Evaluating MAML in our existing setup led to poor performance which we determined was the result of using batch normalization in the base networks.
In the original MAML implementation by \citet{finn2017model}, minibatch statistics are used during both train and test time (that is the batch normalization layers do not track running statistics) and it is assumed evaluation data is also minibatched (see \citep{antoniou2018train} for further details).
In contrast, our setup queries each test example one at a time.
To ensure a fair comparison between all approaches to L2D-Pop and the single-expert baseline, we replaced batch normalization with filter response normalization \citep{singh2020filter} that does not depend on minibatch statistics.
Otherwise the experimental setup is the same as that in \cref{app:image_class_exp_details}.
This leads to lower classifier accuracy as reported in \cref{fig:image_classification_maml_extra}.
There are alternatives developed specifically for meta-learning such as meta-batch normalization \citep{bronskill2020tasknorm} but we could not get it to work for our MAML implementation applied to Pop-L2D.
We leave a more thorough investigation of normalization layers in the case of meta-optimization for Pop-L2D to future work.

MAML for Pop-L2D has two additional hyperparameters, step size and number of steps, for train-time fine-tuning. 
We specify the following grid, step size $[10^{-2}, 10^{-1}]$ and the number of steps $[1,2,5]$.
Due to computational restrictions, we only perform a grid search on these hyperparameters for a single setting of overlap probability $p=0.1$ for each data set (using validation loss as scoring criterion).
The best combination is then used across all $p$.
The only exception is on Traffic Signs for which we observed unstable training for $p \geq 0.4$ and so the hyperparameters were retuned on $p=0.4$.
In summary, the tuned train-time hyperparameters are as follows: for CIFAR-10 and CIFAR-20, we found a step size of $10^{-1}$ and 2 steps was the best; in the case of Traffic Signs, we used a step size of $10^{-1}$ and 5 steps for $p \in (0.1, 0.2)$ and then a reduced step size of $10^{-2}$ for $p \geq 0.4$.
For fine-tuning at evaluation, we used the same step size as during training but allowed the number of steps to be re-tuned (again on the validation loss) on entries in $[1,2,5,10,20]$ greater than or equal to the train-time step count.
We typically observed the selected number of steps at evaluation to be slightly larger than that used during training -- this is similar to what was reported in \citep{finn2017model}.
Similar to \cref{app:image_class_exp_details}, we use vanilla gradient-descent and the base network parameters are frozen during fine-tuning, both train and test-time.
However, in contrast to test-time fine-tuning, we also freeze the classifier parameters (last layer) during train-time fine-tuning (i.e. only fine-tune the rejector).
For the meta-optimization step, we use the implementation of first-order MAML \citep{finn2017model}.

\begin{figure*}[h]
    \begin{subfigure}[t]{.32\linewidth}
        \centering
        \includegraphics{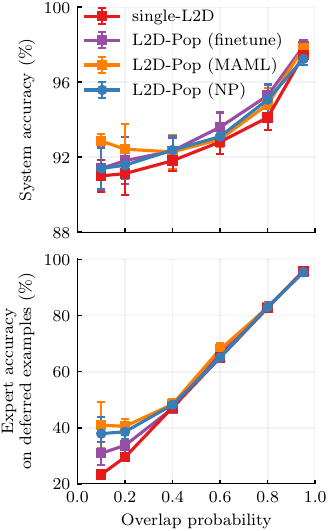}
        \caption{Traffic Signs}
    \end{subfigure}%
    \centering
    \begin{subfigure}[t]{.32\linewidth}
        \centering
        \includegraphics{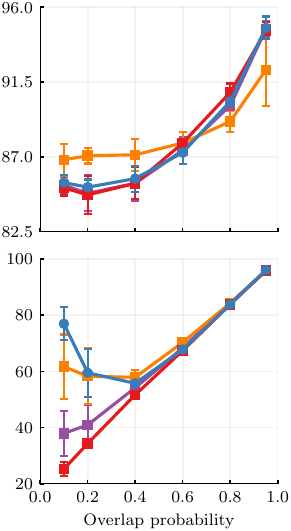}
        \caption{CIFAR-10}
    \end{subfigure}%
    \begin{subfigure}[t]{.32\linewidth}
        \centering
        \includegraphics{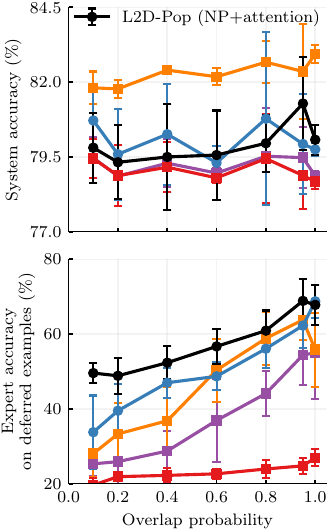}
        \caption{CIFAR-20}
    \end{subfigure}
    \caption{\textit{Varying Population Diversity on Image Classification Tasks} along with MAML approach to L2D-Pop.}
    \label{fig:image_classification_maml}
\end{figure*}

L2D-Pop with MAML along with the other approaches to L2D-Pop and the single-expert baseline are shown in \cref{fig:image_classification_maml} (additional metrics are shown in \cref{fig:image_classification_maml_extra}).
We observe that the MAML implementation consistently improves over test-time only fine-tuning as well as the single-expert baseline in terms of expert accuracy (bottom row). However, the neural process implementation of L2D-Pop remains competitive especially at the lower expert variability setting.
Whilst we observe higher system accuracy for L2D-Pop with MAML, \cref{fig:image_classification_maml_extra} verifies that this gain is largely due to the higher classifier accuracy. 

\begin{algorithm}[t]
    \caption{L2D-Pop with MAML}
    \label{alg:l2d_maml}
    \definecolor{commentgray}{rgb}{0.25, 0.25, 0.25}
    \begin{algorithmic}[1]
        \REQUIRE Step size $\alpha,\beta$, number of inner-optimization steps $S$, $E$ experts with context sets $\{\mathcal{D}_e\}_{e=1}^E$
        \STATE Initialize parameters of classifier $\theta^{\textrm{c}}$ and rejector $\theta^{\textrm{r}}$
        \WHILE{not converged}
            \FOR{expert $e = 1$ \TO $E$}
                \STATE Initialize $\theta_e^{\textrm{r}} = \theta^{\textrm{r}}$
                \FOR{step = $1$ \TO $S$}
                    \STATE {\color{commentgray}\textit{// Compute adapted rejector parameters by gradient descent on expert context set}} %
                    \STATE $\theta_e^{\textrm{r}} \leftarrow \theta_e^{\textrm{r}} - \alpha \nabla_{\theta_e} \sum_{b=1}^B \rphi_{\text{SM}}(\vg^{\theta^{\textrm{c}}}, g_{\bot}^{\theta_e^{\textrm{r}}}; \rd_{e,b})$
                \ENDFOR
            \ENDFOR
            \STATE {\color{commentgray}\textit{// Meta-update using adapted rejector parameters for each expert}}
            \STATE Sample $(\vx,y,\{m_e\}_{e=1}^E)$ from $\mathcal{D}$
            \STATE Update $\theta \leftarrow \theta - \beta \nabla_\theta \rphi_{\text{SM-Pop}}(\vg, g_{\bot}; \vx, y, \{m_e,\theta_e^{\textrm{r}}\}_{e=1}^E)$
        \ENDWHILE
    \end{algorithmic}
\end{algorithm}

\section{ADDITIONAL RESULTS}

\begin{figure*}[h]
    \centering
    \begin{subfigure}[h]{\linewidth}
        \centering
        \includegraphics{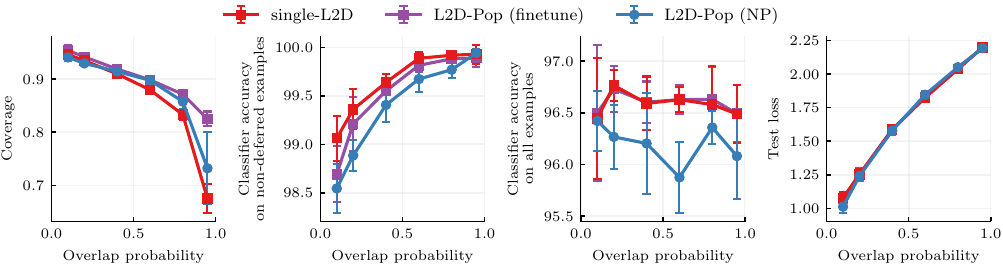}
        \caption{Traffic Signs}
    \end{subfigure}
    \begin{subfigure}[h]{\linewidth}
        \centering
        \includegraphics{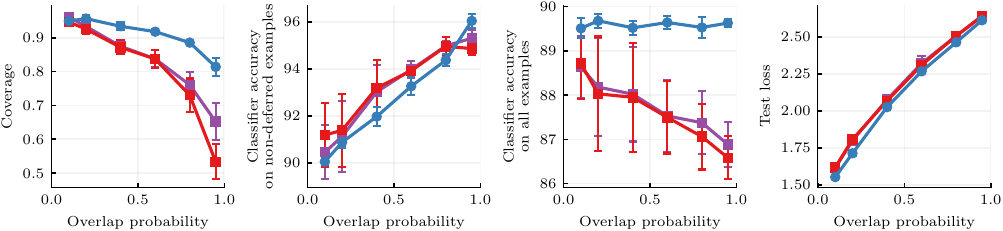}
        \caption{CIFAR-10}
    \end{subfigure}
    \begin{subfigure}[h]{\linewidth}
        \centering
        \includegraphics{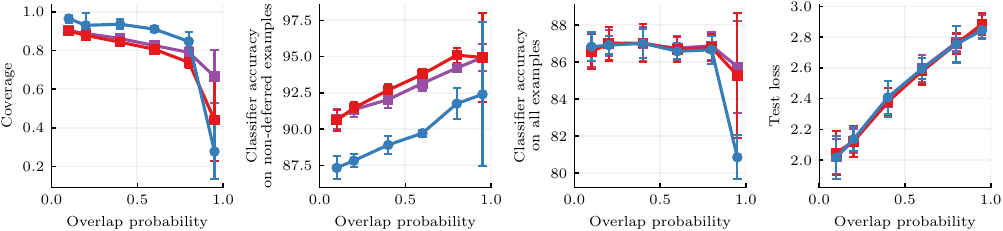}
        \caption{HAM10000}
    \end{subfigure}
    \begin{subfigure}[h]{\linewidth}
        \centering
        \includegraphics{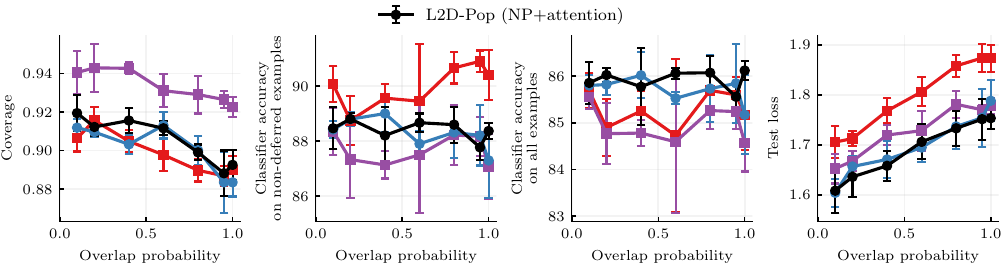}
        \caption{CIFAR-20}
    \end{subfigure}
    \caption{Additional metrics for \textit{varying population diversity on image classification tasks}.}
    \label{fig:image_classification_extra}
\end{figure*}

\begin{figure*}[h]
    \centering
    \begin{subfigure}[h]{\linewidth}
        \centering
        \includegraphics{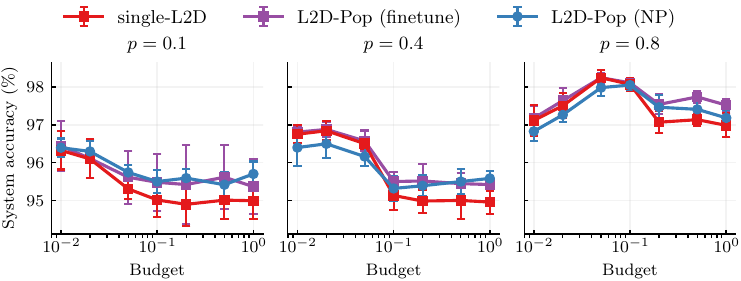}
        \caption{Traffic Signs}
    \end{subfigure}
    \begin{subfigure}[h]{\linewidth}
        \centering
        \includegraphics{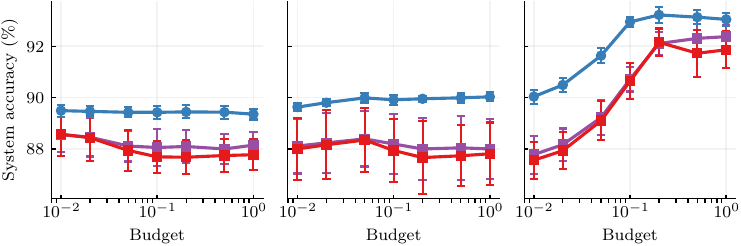}
        \caption{CIFAR-10}
    \end{subfigure}
    \begin{subfigure}[h]{\linewidth}
        \centering
        \includegraphics{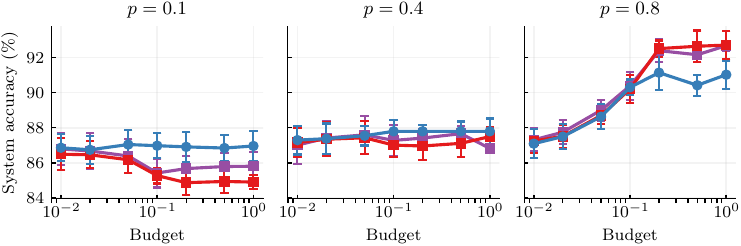}
        \caption{HAM10000}
    \end{subfigure}
    \begin{subfigure}[h]{\linewidth}
        \centering
        \includegraphics{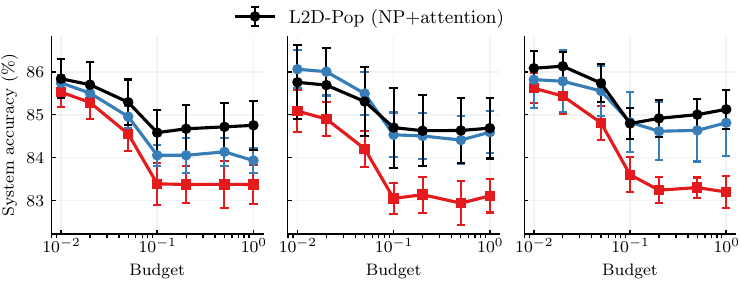}
        \caption{CIFAR-20}
    \end{subfigure}
    \caption{System accuracy as a function of the budget for three settings of the overlap probability $p \in \{0.1,0.4,0.8\}$ corresponding to high, medium and low expert population diversity. This is shown for the data sets and baselines considered in \textit{varying population diversity on image classification tasks}.}
    \label{fig:image_classification_budget}
\end{figure*}

In \cref{fig:image_classification_extra}, we provide additional metrics for the experiments shown in \cref{fig:image_classification,fig:cifar20}.
In \cref{fig:image_classification_budget}, we show the system accuracy as a function of the \emph{budget} for three settings of the overlap probability $p \in \{0.1,0.4,0.8\}$ corresponding to high, medium and low expert population diversity.
The budget is the upper limit on the proportion of examples that can be deferred to the expert.
We refer the reader to App.~E in \citep{verma2022calibrated} for further details as well as details on the implementation.
We observe L2D-Pop by single-expert fine-tuning is competitive against single-expert L2D for a range of budgets considered as well as different diversities of the expert population (not shown in the case of CIFAR-20 due to computational restrictions).
This is also the case for L2D-Pop with neural process except on Traffic Signs and HAM10000 where it is shown to hold for the highest expert population diversity ($p=0.1$) however it is not observed at the lower expert population diversities which show more sensitivity to the budget.

\subsection{One-vs-All (OvA) surrogate}\label{app:exp_ova}
\begin{figure*}[h]
    \begin{subfigure}[t]{.32\linewidth}
        \centering
        \includegraphics{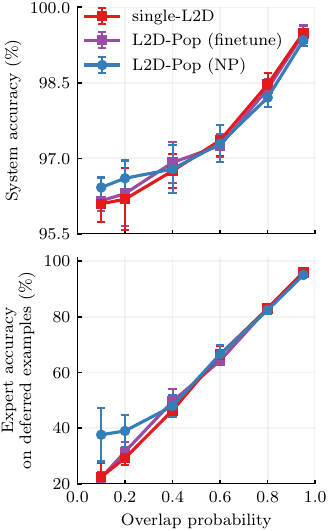}
        \caption{Traffic Signs}
    \end{subfigure}%
    \centering
    \begin{subfigure}[t]{.32\linewidth}
        \centering
        \includegraphics{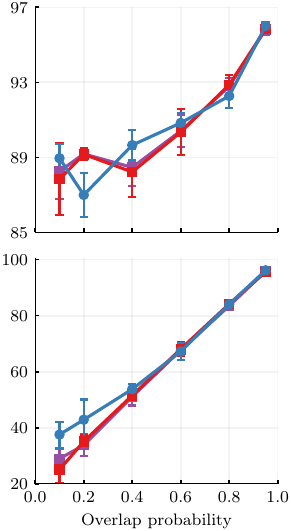}
        \caption{CIFAR-10}
    \end{subfigure}%
    \begin{subfigure}[t]{.32\linewidth}
        \centering
        \includegraphics{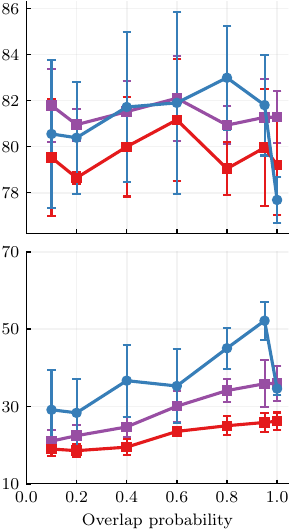}
        \caption{CIFAR-20}
    \end{subfigure}
    \caption{\textit{Varying Population Diversity on Image Classification Tasks} with OvA surrogate loss.}
    \label{fig:image_classification_ova}
\end{figure*}

\begin{figure*}[h]
    \centering
    \begin{subfigure}[h]{\linewidth}
        \centering
        \includegraphics{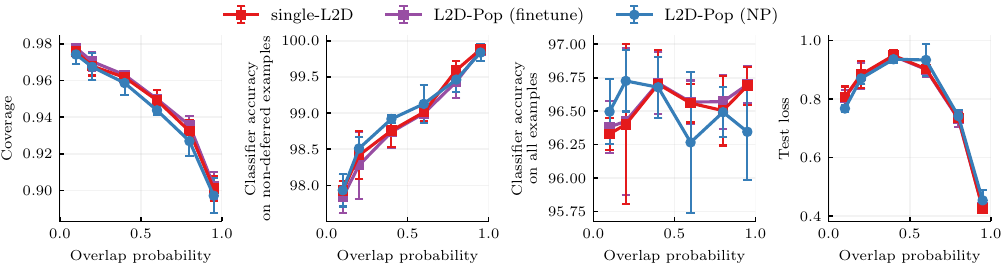}
        \caption{Traffic Signs}
    \end{subfigure}
    \begin{subfigure}[h]{\linewidth}
        \centering
        \includegraphics{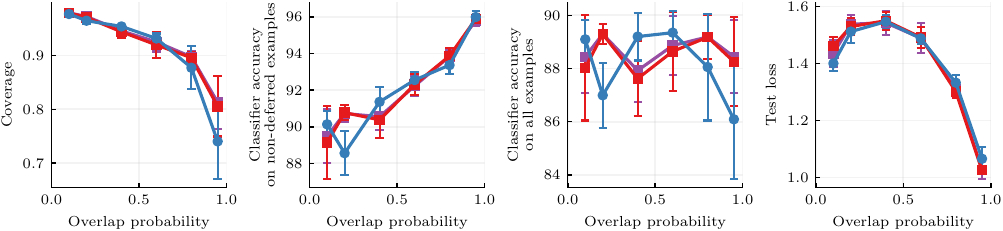}
        \caption{CIFAR-10}
    \end{subfigure}
    \begin{subfigure}[h]{\linewidth}
        \centering
        \includegraphics{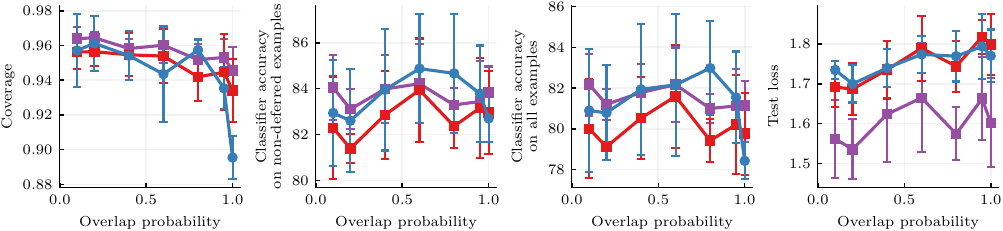}
        \caption{CIFAR-20}
    \end{subfigure}
    \caption{Additional metrics for OvA experiment.}
    \label{fig:image_classification_ova_extra}
\end{figure*}

In \cref{fig:image_classification_ova,fig:image_classification_ova_extra} we perform an ablation where we instead use the OvA surrogate loss for L2D-Pop and the single-expert baseline. All other experimental details are the same as stated in \cref{app:image_class_exp_details} except that a separate base network is trained for the classifier and rejector for all data sets (previously this was only done for CIFAR-20). Similar to the results with the softmax surrogate, we observe an increase in expert accuracy for both fine-tuning and the neural process implementation of L2D-Pop. Except for one setting of overlap probability $p=0.2$ in CIFAR-10, this leads to a boost in system accuracy.

\subsection{Missing context set at test-time}\label{app:missing_context}
\begin{figure*}[h]
    \centering
    \includegraphics[width=0.54\textwidth]{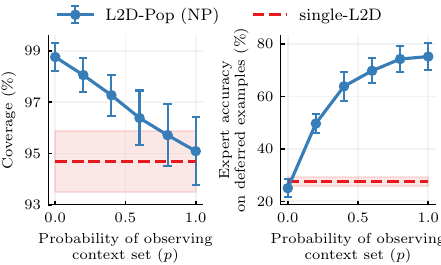}
    \caption{Investigation of the behaviour of L2D-Pop (NP) as the rate at which the context sets are excluded is varied at test-time, from always dropped ($p=0$) to always included ($p=1$).  `Dropping' means that we input only a zero vector. We emphasize that during training of L2D-Pop, context sets are always included ($p=1$). The results are shown on CIFAR-10 for the highest expert diversity setting. Single-expert L2D baseline is shown for comparison.  We see that L2D-Pop simply uses the classifier more and more as context sets are increasingly missing (at test time).}
    \label{fig:missing_context}
\end{figure*}

We investigate the behaviour of the neural process implementation of L2D-Pop when the context set is missing at test time. 
This is done on CIFAR-10 for the highest expert diversity setting: overlap probability of 0.95.
\cref{fig:missing_context} (left) reports the classifier coverage (y-axis) against the probability of observing the context set at test-time.  Thus, at the far left, context sets are never observed, and at the far right, they are always observed.  Our model simply does not defer when the context set is missing: coverage is nearly $99\%$ when the probability is 0.  This is an appropriate behavior since, if the model does not have any information about the available expert, then not deferring is a safe decision.  An example of inappropriate behavior would be making random deferral decisions---which, again, our model is not doing.  

\end{document}

%% file: math_commands.tex
\usepackage{amsmath,amsfonts,bm}
\usepackage{upgreek}

\def\eqref#1{equation~\ref{#1}}

\def\1{\bm{1}}

\def\rphi{{\upphi}}

\def\rd{{\textnormal{d}}}

\def\rsm{{\textnormal{m}}}

\def\rx{{\textnormal{x}}}
\def\ry{{\textnormal{y}}}

\def\rvtheta{{\boldsymbol{\uptheta}}}

\def\rvpsi{{\boldsymbol{\uppsi}}}
\def\rvgamma{{\boldsymbol{\upgamma}}}

\def\rvr{{\mathbf{r}}}

\def\rvx{{\mathbf{x}}}

\def\vmu{{\bm{\mu}}}
\def\vtheta{{\bm{\theta}}}

\def\vpsi{{\bm{\psi}}}

\def\vg{{\bm{g}}}

\def\vx{{\bm{x}}}

\DeclareMathAlphabet{\mathsfit}{\encodingdefault}{\sfdefault}{m}{sl}
\SetMathAlphabet{\mathsfit}{bold}{\encodingdefault}{\sfdefault}{bx}{n}

\DeclareMathOperator*{\argmax}{arg\,max}

\newcommand{\mytick}{\ding{51}}
\newcommand{\mycross}{\ding{53}}